\documentclass[default,iicol]{sn-jnl}

\jyear{2021}%

\theoremstyle{thmstyleone}%
\theoremstyle{thmstyletwo}%

\theoremstyle{thmstylethree}%

\usepackage{graphicx}
\usepackage{amsmath}
\usepackage{amssymb}
\usepackage{booktabs}
\usepackage{xspace}
\usepackage{multirow}

\usepackage[capitalize]{cleveref}
\crefname{section}{Sec.}{Secs.}
\Crefname{section}{Section}{Sections}
\Crefname{table}{Table}{Tables}
\crefname{table}{Tab.}{Tabs.}

\def\onedot{.~}
\def\eg{\emph{e.g}\onedot}

\begin{document}

\title[Article Title]{KBNet: Kernel Basis Network for Image Restoration}








\author[1]{\fnm{Yi} \sur{Zhang}}\email{zhangyi@link.cuhk.edu.hk}
\author[1]{\fnm{Dasong} \sur{Li}}\email{dasongli@link.cuhk.edu.hk}
\author[1]{\fnm{Xiaoyu} \sur{Shi}}\email{xiaoyushi@link.cuhk.edu.hk}
\author[1]{\fnm{Dailan} \sur{He}}\email{hedailan@link.cuhk.edu.hk}
\author[2]{\fnm{Kangning} \sur{Song}}\email{songkangning@sensetime.com}
\author[1]{\fnm{Xiaogang} \sur{Wang}}\email{xgwang@ee.cuhk.edu.hk}
\author[2]{\fnm{Hongwei} \sur{Qin}}\email{qinhongwei@sensetime.com}

\author[1]{\fnm{Hongsheng} \sur{Li}}\email{hsli@ee.cuhk.edu.hk}

\affil[1]{\orgdiv{Multimedia Laboratory}, \orgname{The Chinese University of Hong Kong}}
\affil[2]{\orgdiv{SenseTime Research}}


\abstract{

How to aggregate spatial information plays an essential role in learning-based image restoration.
Most existing CNN-based networks adopt static convolutional kernels to encode spatial information, which cannot aggregate spatial information adaptively. 
Recent transformer-based architectures achieve adaptive spatial aggregation. But they lack desirable inductive biases of convolutions and require heavy computational costs. 
In this paper, we propose a kernel basis attention (KBA) module, which introduces learnable kernel bases to model representative image patterns for spatial information aggregation. 
Different kernel bases are trained to model different local structures. 
At each spatial location, they are linearly and adaptively fused by predicted pixel-wise coefficients to obtain aggregation weights.
Based on the KBA module, we further design a multi-axis feature fusion (MFF) block to encode and fuse channel-wise, spatial-invariant, and pixel-adaptive features for image restoration.
Our model, named kernel basis network (KBNet), achieves state-of-the-art performances on more than ten benchmarks over image denoising, deraining, and deblurring tasks while requiring less computational cost than previous SOTA methods. Code will be released at \url{https://github.com/zhangyi-3/kbnet}.

}




\maketitle

\section{Introduction}
\label{sec:intro}
Image restoration is one of the most foundational tasks in computer vision, which aims to remove the unavoidable degradation of the input images and produce clean outputs. Image restoration is highly challenging as it is an ill-posed problem. It does not only play an important role in a wide range of low-level applications (\eg night sight on smartphones) but also benefits many high-level vision tasks~\cite{LiuWLWH18}.

Linearly aggregating spatial neighborhood information is a common component and plays a key role in deep neural networks for feature encoding in image restoration.
Convolutional neural networks (CNNs) are one of the dominant choices for local information aggregation. 
They use globally-shared convolution kernels in the convolution operator for aggregating neighboring information, such as using dilated convolutions~\cite{RIDNet, chang2020sadnet} to increase the receptive fields and adopting multi-stage~\cite{Zamir2021MPRNet} or multi-scale features~\cite{gu2019self,zamir2020mirnet} for better encoding spatial context.
While CNN-based methods show clear performance gains than traditional handcrafted methods~\cite{BM3D,NLM,wavelet,oldksvd}, 
the convolutions utilized static and spatially invariant kernels for all spatial locations and therefore have limited capacity to handle different types of image structures and textures.
While a few works~\cite{kpn,xia2020bpn,mckpn} in the low-level task of burst image denoising were proposed to adaptively encode local neighborhoods features,
they require heavy computational costs to predict adaptive convolutional kernels of each output pixel location.

Recently, the vision transformers have shown great progress where the attention mechanism uses dot products between pairwise positions to obtain the linear aggregation weights for encoding local information. 
A few efforts~\cite{chen2021IPT,wang2021uformer,liang2021swinir,tu2022maxim} have been made in image restoration. In IPT~\cite{chen2021IPT}, the global attention layers are adopted to aggregate from all spatial positions, which, however, face challenges on handling high-resolution images.
A series of window-based self-attention solutions~\cite{wang2021uformer, liang2021swinir, tu2022maxim} have been proposed to alleviate the quadratic computational cost with respect to the input size. 
While self-attention enables each pixel to adaptively aggerate spatial information from the assigned window, it lacks the inductive biases possessed by convolutions (\eg locality, translation equivalence, etc.), which is useful in modeling local structures of images.
Some work~\cite{liu2022convnext,han2021connection,replknet} also reported that using convolutions to aggregate spatial information can produce more satisfying results than self-attention.

In this paper, to tackle the challenge of effectively and efficiently aggregating neighborhood information for image restoration, we propose a kernel basis network (KBNet) with a novel kernel basis attention (KBA) module to adaptively aggregate spatial neighborhood information, which takes advantage of both CNNs and transformers.
Since natural images generally share similar patterns across different spatial locations, 
we first introduce a set of learnable kernel bases to learn various local patterns.
The kernel bases specify how the neighborhood information should be aggregated for each pixel, and different learnable kernel bases are trained to capture various image patterns.
Intuitively, similar local neighborhoods would use similar kernel bases so that the bases are learned to capture more representative spatial patterns.
Given the kernel bases, we adopt a separate lightweight convolution branch to predict the linear combination coefficients of kernel bases for each pixel. 
The learnable kernel bases are linearly fused by the coefficients to produce adaptive and diverse aggregation weights for each pixel. 

Unlike the window-based self-attention in previous works~\cite{liang2021swinir,wang2021uformer,tu2022maxim} that uses dot products between pairwise positions to generate spatial aggregation weights, the KBA module generates the weights by adaptively combining the learnable kernel bases. 
It naturally keeps the inductive biases of convolutions while handling various spatial contexts adaptively. 
KBA module is also different from existing dynamic convolutions~\cite{dynamicConvGao,malleconv} or kernel prediction networks~\cite{mckpn,kpn,xia2020bpn,Wang2019Carafe}, which predict all the kernel weights directly. The aggregation weights by the KBA module are predicted by spatially adaptive fusing the shared kernel bases. Thus, our KBA module is more lightweight and easier to optimize. The ablation study validates the proposed design choice. 

Based on the KBA module, we design a multi-axis feature fusion (MFF) block to extract and fuse diverse features for image restoration.
We combine operators of spatial and channel dimensions to better capture image context.
In the MFF block, three branches, including channel attention, spatial-invariant, and pixel-wise adaptive feature extractions are parallelly performed and then fused by point-wise production.

By integrating the proposed KBA module and MFF block into the widely used U-Net architectures with two variants of feed-forward networks, our proposed KBNet achieves state-of-the-art performances on more than ten benchmarks of image denoising, deraining, and deblurring.

The main contributions of this work are summarized as follows:
\begin{enumerate}
    \item We propose a novel kernel basis attention (KBA) module to effectively aggregate the spatial information via a series of learnable kernel bases and linearly fusing the kernel bases.
    \item We propose an effective multi-axis feature fusion (MFF) block, which extracts and fuses channel-wise, spatial-invariant, and pixel-wise adaptive features for image restoration.
    \item Our method kernel basis network (KBNet) demonstrates its generalizability and state-of-the-art performances on three image restoration tasks including denoising, deblurring, and deraining.
\end{enumerate}

\section{Related Work}
Deep image restoration has been a popular topic in the field of computer vision and has been extensively studied for many years. In this section, we review some of the most relevant works. 

\subsection{Traditional Methods}
Since image restoration is a highly ill-posed problem, many priors or noise models~\cite{zhang2021rethinkingNoise} are adopted to help image restoration. 
Many properties of natural images have been discussed in traditional methods, like sparsity~\cite{mairal2007sparse,SparseDictionary}, non-local self-similarity~\cite{BM3D,NLM}, total variation~\cite{variationNoiseRemove}.
Self-similarity is one of the natural image properties used in image restoration. Traditional methods like non-local means~\cite{NLM}, BM3D~\cite{BM3D} leverage the self-similarity to denoise images by averaging the intensities of similar patches in the image. 
While traditional methods have achieved good performance, they tend to produce blurry results and fail in more challenging cases. 

\subsection{CNNs for Image Restoration}
\noindent\textbf{Static networks}.~
With the popularity of deep neural networks, learning-based methods become the mainstream of image restoration~\cite{zhang2020rdn,zamir2020mirnet,Zamir2021MPRNet,chen2022nafnet,cho2021rethinking_mimo,dasongBlurRepresentaion,GroupShift,ntire2019_denoising,ntire2019_superresolution,nah2021ntire}.
One of the earliest works on deep image denoising using CNNs is the DnCNN~\cite{DnCNN}. Then, lots of CNN-based models have been proposed from different design perspectives:
residual learning~\cite{RIDNet,RCAN}, multi-scale features~\cite{zamir2020mirnet}, multi-stage design~\cite{Zamir2021MPRNet,BurstRawVS}, non-local information~\cite{zhang2019residual,plotz2018N3Net}.
While they improve the learning capacity of CNN models, they use the normal convolutions as the basic components, which are static and spatially invariant. As a result, plain and textural areas cannot be identified and processed adaptively, which is crucial for image restoration.

\noindent\textbf{Dynamic networks for Image Restoration}.~
Another line of research has focused on leveraging dynamic networks to image restoration tasks.
Kernel prediction networks~\cite{kpn,mckpn, xia2020bpn,malleconv} (KPNs) use the kernel-based attention mechanism in a CNN architecture to aggregate spatial information adaptively. But, KPNs predict the kernels directly which have significant memory and computing requirements, as well as their difficulty to optimize.
\subsection{Transformers for Image Restoration}
Transformers have shown great progress in natural language and high-level vision tasks.
For image restoration, IPT~\cite{chen2021IPT} is the first method to adopt a transformer (both encoder and decoder) into image restoration. But, it leads to heavy computational costs and can only perform on fixed patch size $48 \times 48$. Most of the following works only utilize the encoder and tend to reduce the computational cost. A seris of window-based self-attention~\cite{liang2021swinir,wang2021uformer,tu2022maxim} has been proposed. 
Each pixel can aggregate the aggregate spatial information through the dot-productions between all pairwise positions.
More recently, some work~\cite{liu2022convnext,han2021connection,replknet,chen2022nafnet} indicate that self-attention is not necessary to achieve state-of-the-art results.
  
\begin{figure*}[t]
    \centering
    \includegraphics[width=0.95\linewidth]{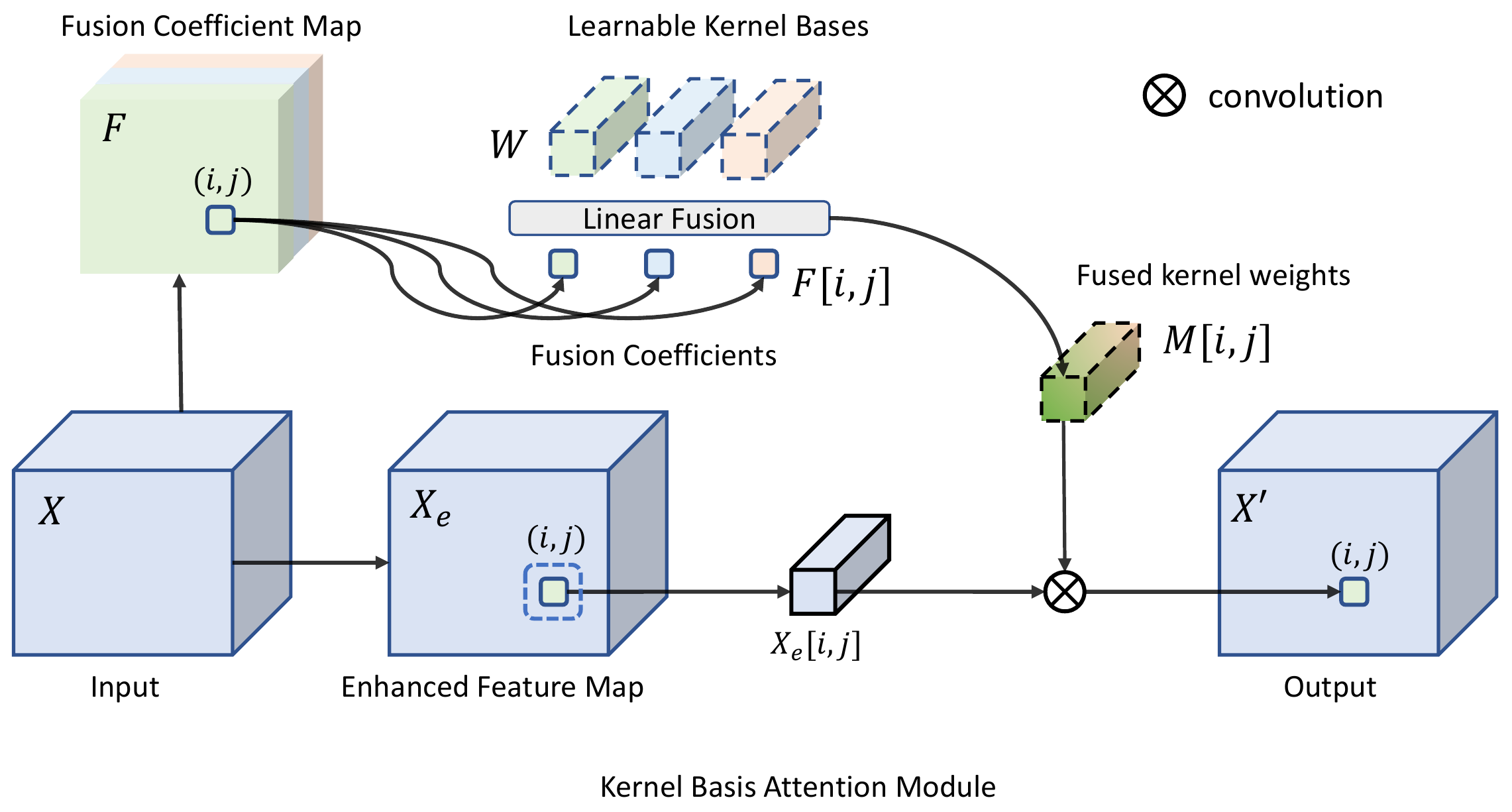}
    \caption{An overview of kernel basis attention (KBA) Module. With the input feature map $X$, the KBA module first predicts the fusion coefficient map $F$ to linearly fuse the learnable kernel bases $W$ for each location. Then, the fused kernel weights $M$ adaptively encode the local neighborhood of the enhanced feature map $X_e$ to produce the output feature map $X'$.}
    \label{fig:overview}
\end{figure*}

\section{Method}
In this section, we aim to develop a novel kernel basis network (KBNet) for image restoration.
We first describe the kernel basis attention (KBA) module to adaptively aggregate the spatial information. Then, the multi-axis feature fusion (MFF) block is introduced to encode and fuse diverse features for image restoration.
Finally, we describe the integration of the MFF block into the U-Net.

\subsection{Kernel Basis Attention Module}
How to gather spatial information at each pixel plays an essential role in feature encoding for low-level vision tasks. Most CNN-based methods~\cite {DnCNN,cheng2021nbnet,RIDNet} utilize spatial-invariant kernels to encode spatial information of the input image, which cannot adaptively process local spatial context for each pixel.
While self-attention~\cite{tu2022maxim,wang2021uformer,liang2021swinir} 
 can process spatial context adaptively according to the attention weights from the dot products between pairwise positions, it lacks the inherent inductive biases of convolutions.
To tackle the challenges, 
we propose the kernel basis attention (KBA) module to encode spatial information by fusing learnable kernel bases adaptively.

As shown in \cref{fig:overview}, given an input feature map $X \in R^{H \times W \times C}$, our KBA learns a set of learnable kernel bases $W$ shared across all spatial locations and all images to capture common spatial patterns.
The learnable kernel bases $W \in \mathbb{R}^{N \times C \times 4 \times K^2}$ contains $N$ grouped convolution kernels.
The $C$ and $K^2$ denote channel number and kernel size respectively. The group size is set to $\frac{C}{4}$ for balancing the performance-efficiency tradeoff.

\noindent\textbf{Fusion Coefficients Prediction}.~
To adaptively combine the $N$ learnable kernel bases at each spatial location, given the input feature map $X \in \mathbb{R}^{H \times W \times C}$, a lightweight convolution branch is used to predict the $N$ kernel bases fusion coefficients $F \in \mathbb{R}^{H \times W \times N}$ at each location.
The lightweight convolution branch contains two layers.
The first $3\times3$ grouped convolution layer reduces the feature map channel to $N$ with group size $N$. 
To further transform the features, a SimpleGate activation function~\cite{chen2022nafnet} followed by another $3\times3$ convolution layer is adopted.

Here, a natural choice is to normalize the fusion coefficients at each location by the softmax function so that the fusion coefficients sum up to 1. However, we experimentally find that using softmax normalization hinders the final performance since it tends to select the most important kernel basis instead of fusing multiple kernel bases for capturing spatial contexts.

\noindent\textbf{Kernel Basis Fusion}.~
With the predicted fusion coefficient map $F \in \mathbb{R}^{H \times W \times N}$ and kernel bases $W \in \mathbb{R}^{N \times C \times 4 \times K^2}$, the fused weights $M[i, j]$ for the spatial position $(i, j)$ can be obtained by the linear combination of learnable kernel bases:
\begin{align*}
    M[i, j] = \sum_{t=0}^{N} F[i, j, t]  W[t],
\end{align*}
    where $W[t] \in \mathbb{R}^{C \times 4 \times K^2}$ denotes the $t$-th learnable kernel basis, and $F[i, j, t] \in \mathbb{R}^{N}$ and $M[i, j] \in \mathbb{R}^{C \times 4 \times K^2}$ are the $t$-th kernel fusion coefficients and the fused kernel weights at the position $(i, j)$ respectively. 
Besides, the input feature map $X$ is transformed by a $1 \times 1$ 
 convolution to obtain the feature map $X_e$ for adaptive convolution with the fused kernel weights $M$.
The output feature map $X'$ at position $(i, j)$ 
is therefore obtained as
\begin{align*}
    X'[i, j] = \mathrm{GroupConv}(X_e[i, j], M[i, j]),
\end{align*}
where $X' \in \mathbb{R}^{H \times W \times C}$ is the output feature map and maintains the input spatial resolution.

\noindent\textbf{Discussion}.~
Previous kernel prediction methods~\cite{kpn,mckpn,malleconv,xia2020bpn} also predict pixel-wise kernels for convolution. But they adopt a heavy design to predict all kernel weights directly.
Specifically, even predicting $K \times K$ depthwise kernels requires producing a $(C \times K \times K)$ channel feature map, which is quite costly in terms of both computational cost and memory. In contrast, Our method trains a series of learnable kernel bases, which only needs to predict an $N$-channel fusing coefficient map (where $N \ll C \times K^2$).
Such a design avoids predicting a large number of kernel weights for each pixel and the representative kernel bases shared across all locations are still efficiently optimized.
Compared with window-based self-attention~\cite{wang2021uformer,tu2022maxim}, our spatial aggregation weights are linearly combined from the shared kernel bases instead of being produced individually through dot products between pairwise positions. 
The KBA module adopts a set of learnable convolution kernels for modeling different local structures and fuses those kernel weights adaptively for each location.
Thus, it benefits from the inductive bias of convolutions while achieving spatially-adaptive processing effectively.

\begin{figure}[t]
    \centering
    \includegraphics[width=0.95\linewidth]{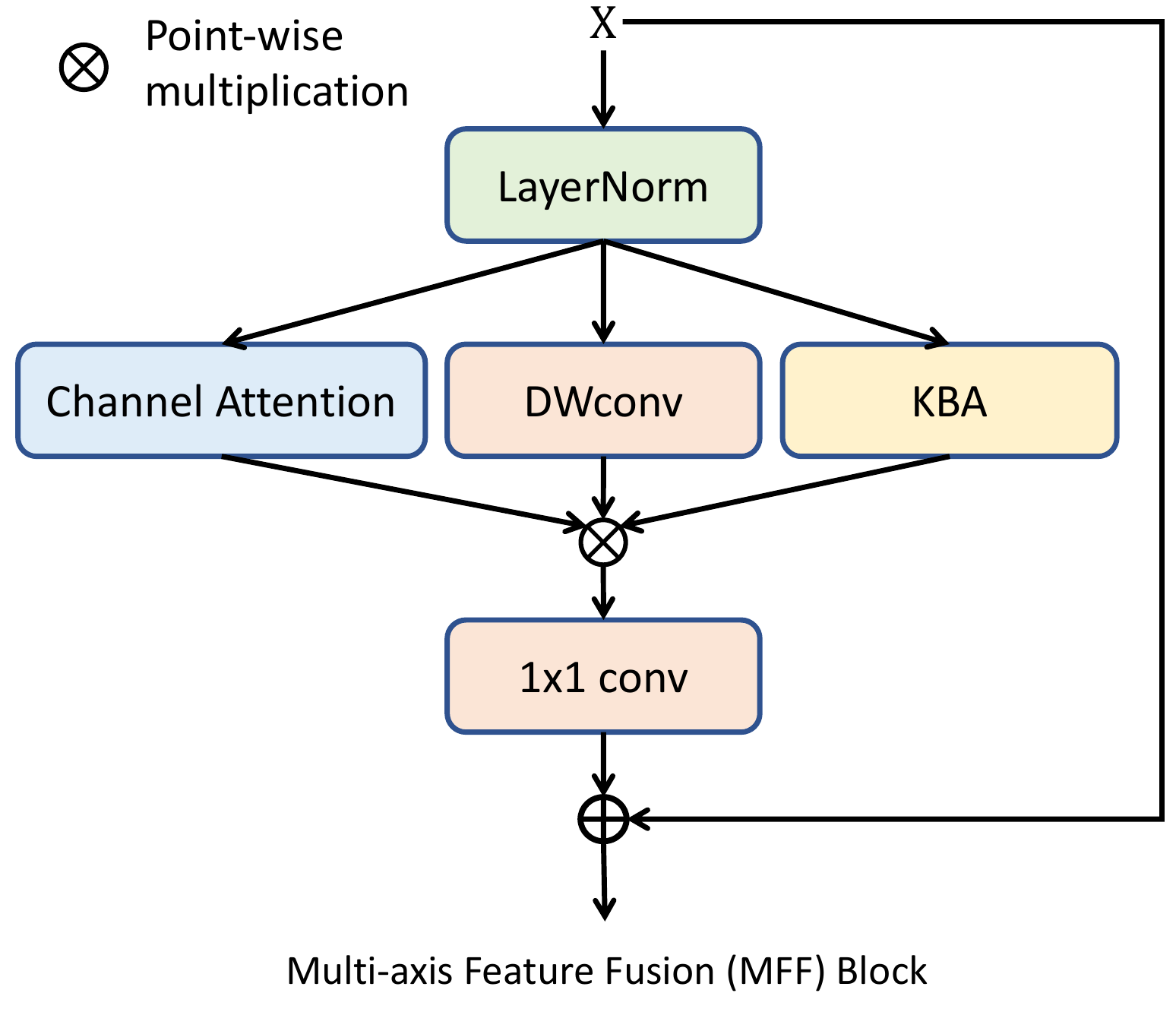}
    \caption{An overview of Multi-axis Feature Fusion (MFF) Block. Channel attention, depthwise convolution, and our KBA module process the input features parallelly. The outputs of three operations are fused by point-wise multiplication. }
    \label{fig:block}
\end{figure}

\renewcommand{\arraystretch}{1.} 

\begin{table*}[t]
    \centering
    \caption{Denoising results of color images with Gaussian noise on four testing datasets and three noise levels. The \textbf{best results} are marked by bold fonts.}
\setlength{\tabcolsep}{2pt}
\scalebox{0.85}{
\begin{tabular}{l | c c c | c c c | c c c | c c c | c}
\toprule
   & \multicolumn{3}{c|}{ \textbf{CBSD68}~\cite{martin2001database_bsd}} & \multicolumn{3}{c|}{ \textbf{Kodak24}~\cite{kodak}} & \multicolumn{3}{c|}{ \textbf{McMaster}~\cite{zhang2011color_mcmaster}} & \multicolumn{3}{c|}{ \textbf{Urban100}~\cite{huang2015single_urban100}} \\
 \cline{2-13}
    \textbf{Method} & $\sigma$$=$$15$ & $\sigma$$=$$25$ & $\sigma$$=$$50$ & $\sigma$$=$$15$ & $\sigma$$=$$25$ & $\sigma$$=$$50$ & $\sigma$$=$$15$ & $\sigma$$=$$25$ & $\sigma$$=$$50$ & $\sigma$$=$$15$ & $\sigma$$=$$25$ & $\sigma$$=$$50$ & \textbf{MACs}\\
\midrule
IRCNN~\cite{zhang2017learning}   & 33.86 & 31.16 & 27.86 & 34.69 & 32.18 & 28.93 & 34.58 & 32.18 & 28.91 & 33.78 & 31.20 & 27.70&-\\
FFDNet~\cite{FFDNetPlus}  &33.87 & 31.21 & 27.96 & 34.63 & 32.13 & 28.98 & 34.66 & 32.35 & 29.18 & 33.83 & 31.40 & 28.05 &-\\
DnCNN~\cite{DnCNN}  &33.90 & 31.24 & 27.95 & 34.60 & 32.14 & 28.95 & 33.45 & 31.52 & 28.62 & 32.98 & 30.81 & 27.59 &37G\\
DSNet~\cite{peng2019dilated}  & 33.91 & 31.28 & 28.05 & 34.63 & 32.16 & 29.05 & 34.67 & 32.40 & 29.28 & - & - & -\\
DRUNet~ &{34.30} &  {31.69} &  {28.51} &  {35.31} &  {32.89} &  {29.86} &  {35.40} &  {33.14} &  {30.08} &  {34.81} &  {32.60} &  {29.61} & 144G\\
RPCNN~\cite{xia2020rpcnn}  &- & 31.24 & 28.06 & - & 32.34 & 29.25 & - & 32.33 & 29.33 & - & 31.81 & 28.62&-\\
BRDNet~\cite{tian2020BRDnet}  &34.10 & 31.43 & 28.16 & 34.88 & 32.41 & 29.22 & 35.08 & 32.75 & 29.52 & 34.42 & 31.99 & 28.56 &-\\
RNAN~\cite{zhang2019residual}  &-&-&28.27&-&-&29.58&-&-&29.72&-&-&29.08 &496G\\
RDN~\cite{zhang2020rdn}  &-&-&28.31&-&-&29.66&-&-&-&-&-&29.38& 1.4T\\  
IPT~\cite{chen2021IPT}  &-&-&28.39&-&-&29.64&-&-&29.98&-&-&29.71 & 512G\\
SwinIR~\cite{liang2021swinir} &  {34.42} &  {31.78} &  {28.56} &  {35.34} &  {32.89} &  {29.79} &  {35.61} &  {33.20} &  {30.22} &  {35.13} &  {32.90} &  {29.82} & 759G\\
Restormer~\cite{restormer} &  {34.40} &  {31.79}&  {28.60}&  \textbf{35.47} &  {33.04}&  {30.01}&  \textbf{35.61}&  \textbf{33.34}&  \textbf{30.30} &  {35.13}&  {32.96}&  {30.02} & 141G\\ 
 $\mathrm{\textbf{KBNet}}_s$ & \textbf{34.41} & \textbf{31.80} & \textbf{28.62} & 35.46 & \textbf{33.05} & \textbf{30.04} & 35.56 & 33.31 & 30.27 & \textbf{35.15}  & \textbf{32.96} &  \textbf{30.04} & 69G\\
\bottomrule

\end{tabular}}
    
    \label{tab:c_gaussian}
\end{table*}
\renewcommand{\arraystretch}{1.} 

\subsection{Multi-axis Feature Fusion Block}
To encode diverse features for image restoration, based on the KBA module, we design a Multi-axis Features Fusion (MFF) block to handle channel and spatial information. 
As shown in \cref{fig:block}, MFF block first adopts a layer normalization to stabilize the training and then performs spatial information aggregation.  A residual shortcut is used to facilitate training convergence.
Following the normalization layer, three operators are adopted in parallel. The first operator is a $3 \times 3$ depthwise convolution to capture spatially-invariant features. The second operator is channel attention~\cite{hu2019squeeze} to modulate the feature channels. The third one is our KBA module to adaptively handle spatial features.
The three branches output feature maps of the same size.
Point-wise multiplication is used to fuse the diverse feature from the three branches directly, which also serves as the non-linear activation~\cite{chen2022nafnet} of the MFF block.

\subsection{Intergration of MFF Block into U-Net}
We adopt the widely used U-shape architectures.
The U-shape architecture processes the input noisy image $I \in \mathbb{R}^{H \times W \times 3}$ to generate a clean image of the same resolution. 
The first convolution layer transforms the input image into the feature map $F \in \mathbb{R}^{H \times W \times C}$. 
Then, the feature map $F$ is processed by an encoder-decoder architecture, each of which has four stages. 
At each encoder stage, the input resolution is downsampled by a convolution layer with stride 2 and the channels are expanded by a factor of 2. 
Within each stage, the MFF blocks are stacked in both the encoder and decoder.
In each decoder stage, the feature map resolution and the channels are reversed by the pixel-shuffle operation. 
Besides, the shortcuts from the encoder are passed to the decoder layer of the same stages and fused simply by an addition operation. 
The last convolution transforms the feature map to the same shape as the input.
Inspired by the design of transformer blocks, the MFF block is also followed by a Feed Forward Network (FFN) at each stage. 
We present two variants of our method $\mathrm{KBNet}_s$ and $\mathrm{KBNet}_l$.
$\mathrm{KBNet}_s$ adopts the normal FFN block with SimpleGate~\cite{chen2022nafnet} activation function to perform the position-wise non-linear transformation.
$\mathrm{KBNet}_l$ utilizes a heavier transposed attention in Restormer~\cite{restormer} to achieve more complex position-wise transformation.
The computational costs of $\mathrm{KBNet}_s$ and $\mathrm{KBNet}_l$ are around 70G and 108G MACs separately. Detailed network architectures can be found in \cref{sec:implementation}.

\section{Results}
In this section, we first describe the implementation details of our methods. Then, we evaluate our KBNet on popular benchmarks over synthetic denoising, real-world denoising, deraining and deblurring tasks.
Finally, we conduct ablation studies on Gaussian denoising to validate important designs of our method and compare KBNet with existing methods.

\begin{figure}[t]
    \centering
    \includegraphics[width=0.95\linewidth]{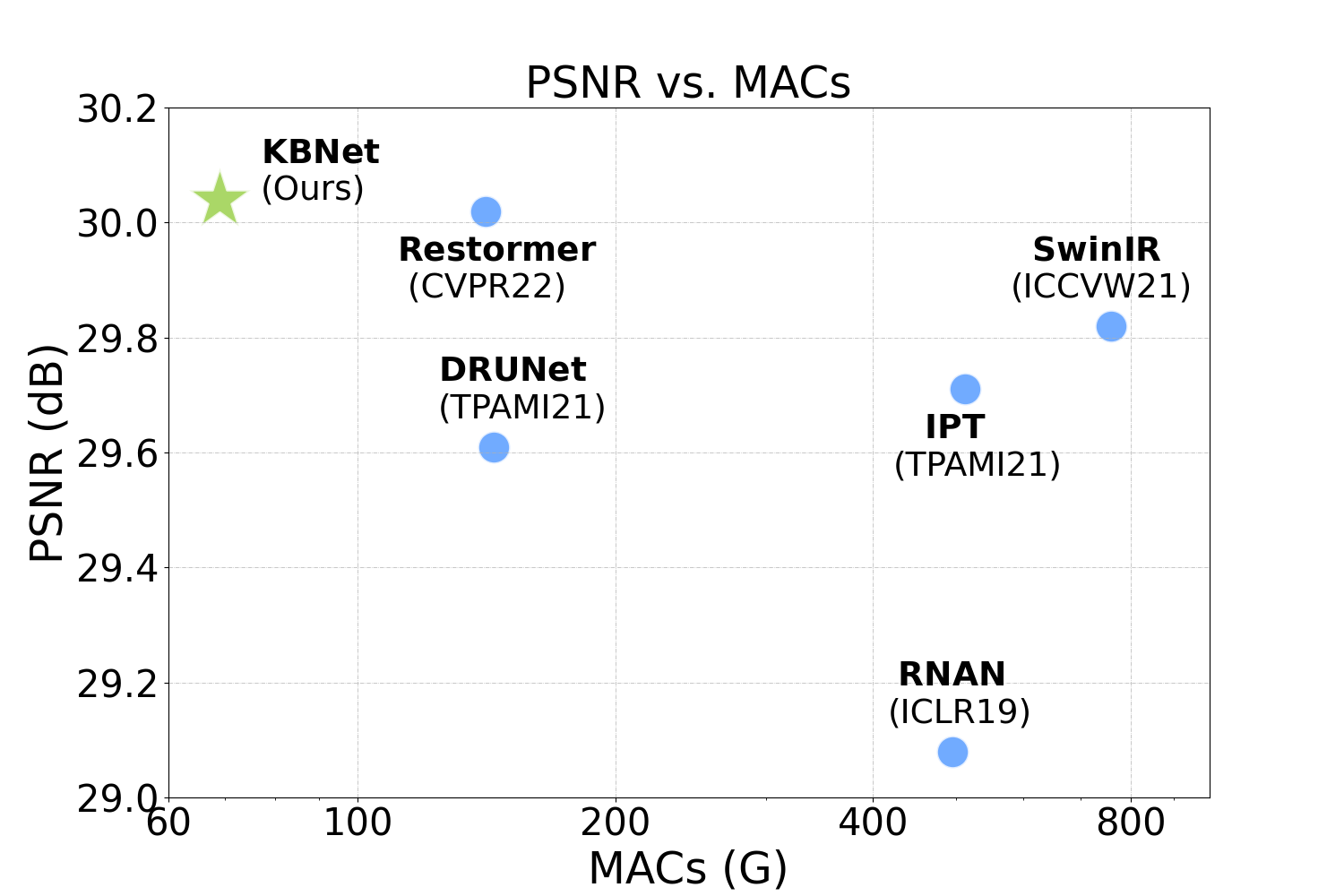}
    \caption{PSNR v.s  MACs of different methods on Gaussian denoising of color images. PSNRs are tested on Urban dataset with noise level $\sigma=50$.}
    \label{fig:trade_g}
\end{figure}

\subsection{Implementation Details}\label{sec:implementation}
The default setting of our method is introduced as follows unless otherwise specified. The block numbers for each stage in the encoder and decoder of our KBNet are $\{2, 2, 2, 2\}$ and $\{4, 2, 2, 2\}$, respectively.
For the KBA module, The number of kernel basis is set to 32 by default. For each kernel base, we use a grouped convolution kernel with kernel size $3 \times 3$ and 4 channels for each group.
 We train our model for 300k iterations for each noise level. The patch size is $256 \times 256$, batch size is 32, and learning rate is $10^{-3}$ following the training settings of NAFNet~\cite{chen2022nafnet}.

\renewcommand{\arraystretch}{1.} 

\begin{table*}
\centering
\caption{Gaussian denoising results of gray images on three testing datasets and three noise levels. The \textbf{best result} are marked in bold.}

\scalebox{0.9}{
\begin{tabular}{l | c c c | c c c | c c c| c}
\toprule
   & \multicolumn{3}{c|}{ \textbf{Set12}~\cite{DnCNN}} & \multicolumn{3}{c|}{ \textbf{BSD68}~\cite{martin2001database_bsd}} & \multicolumn{3}{c|}{ \textbf{Urban100}~\cite{huang2015single_urban100}} & \\
 \cline{2-10}
    \textbf{Method} & $\sigma$$=$$15$ & $\sigma$$=$$25$ & $\sigma$$=$$50$ & $\sigma$$=$$15$ & $\sigma$$=$$25$ & $\sigma$$=$$50$ & $\sigma$$=$$15$ & $\sigma$$=$$25$ & $\sigma$$=$$50$ & \textbf{MACs} \\
\midrule
DnCNN~\cite{DnCNN}  &32.67 & 30.35 & 27.18 & 31.62 & 29.16 & 26.23 & 32.28 & 29.80 & 26.35&37G\\
FFDNet~\cite{FFDNetPlus}  &32.75 & 30.43 & 27.32 & 31.63 & 29.19 & 26.29 & 32.40 & 29.90 & 26.50&-\\ 
IRCNN~\cite{zhang2017learning}  &32.76 & 30.37 & 27.12 & 31.63 & 29.15 & 26.19 & 32.46 & 29.80 & 26.22&-\\ 
DRUNet~\cite{zhang2021DPIR}  &  {33.25} &  {30.94} &  {27.90} &  {31.91} &  {29.48} &  {26.59} &  {33.44} &  {31.11} &  {27.96}&144G\\ 
FOCNet~\cite{jia2019focnet}  &33.07 & 30.73 & 27.68 & 31.83 & 29.38 & 26.50 & 33.15 & 30.64 & 27.40&-\\
MWCNN~\cite{liu2018MWCNN}  &33.15 & 30.79 & 27.74 & 31.86 & 29.41 & 26.53 & 33.17 & 30.66 & 27.42&-\\
NLRN~\cite{liu2018NLRN}  &33.16 & 30.80 & 27.64 & 31.88 & 29.41 & 26.47 & 33.45 & 30.94 & 27.49&-\\
RNAN~\cite{zhang2019residual}  &- & - & 27.70 & - & - & 26.48 & - & - & 27.65 & 496G\\
DeamNet~\cite{ren2021adaptivedeamnet}  &33.19 & 30.81 & 27.74 & 31.91 & 29.44 & 26.54 & 33.37 & 30.85 & 27.53 &146G\\
DAGL~\cite{mou2021dynamicDAGL}  &33.28 & 30.93 & 27.81 & 31.93 & 29.46 & 26.51 & 33.79 & 31.39 & 27.97 &256G\\
SwinIR~\cite{liang2021swinir} &  {33.36} &  {31.01} &  {27.91} &  {31.97} &  {29.50} &  {26.58} &  {33.70} &  {31.30} &  {27.98} & 759G\\ 
Restormer~\cite{restormer} &  \textbf{33.42} &  {31.08} &  {28.00} &  {31.96} &  {29.52}&  {26.62}&  \textbf{33.79} &  \textbf{31.46}&  {28.29} & 141G\\
$\mathrm{\textbf{KBNet}}_s$ & 33.40 & \textbf{31.08} & \textbf{28.04} & \textbf{31.98} & \textbf{29.54} & \textbf{26.65} & 33.77 & 31.45 & \textbf{28.33} & 69G
\\
\bottomrule
\end{tabular}}

    \label{tab:g_gaussian}
\end{table*}
\renewcommand{\arraystretch}{1.} 

\setlength{\tabcolsep}{1pt}
\def \imgl {0.14\linewidth}
\def \imgs {0.14\linewidth}
\begin{figure*}[!t]
    \small
    \begin{center}
    \begin{tabular}{cccccc}
        \includegraphics[width=\imgl]{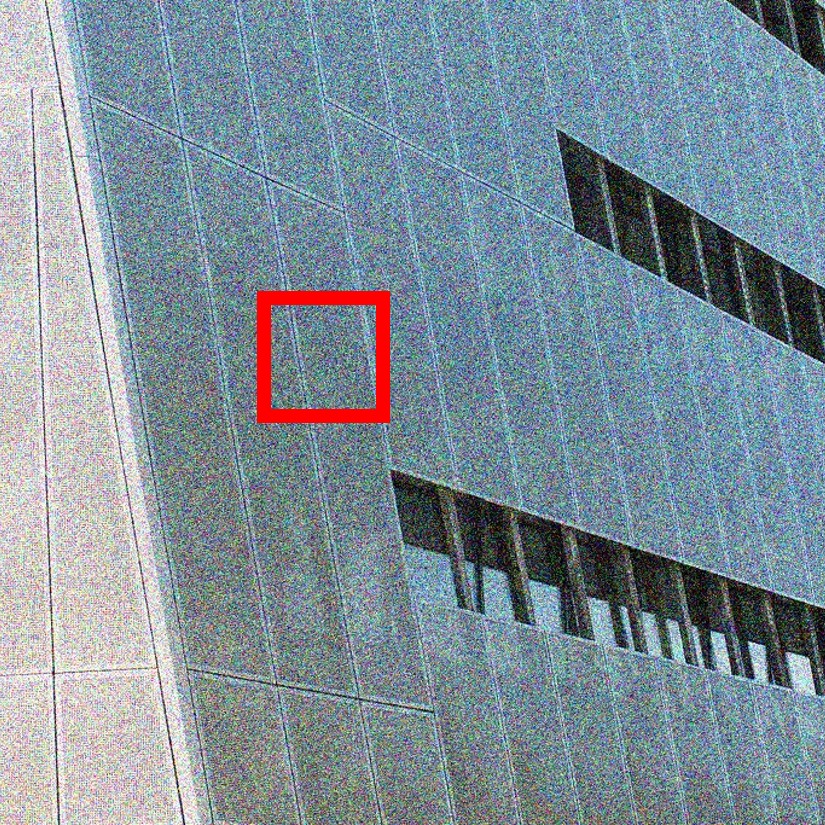} &
        \includegraphics[width=\imgs]{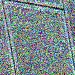} &
        \includegraphics[width=\imgs]{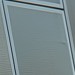} &
        \includegraphics[width=\imgs]{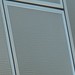} &
        \includegraphics[width=\imgs]{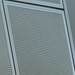} &
        \includegraphics[width=\imgs]{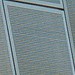} \\
        \includegraphics[width=\imgl]{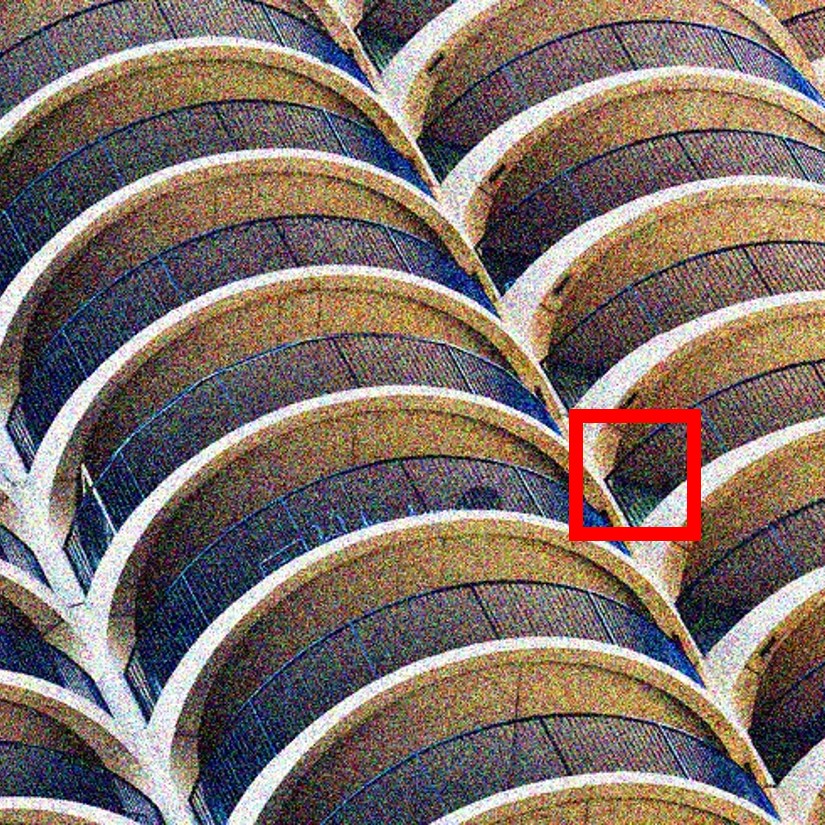} &
        \includegraphics[width=\imgs]{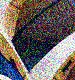} &
        \includegraphics[width=\imgs]{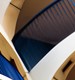} &
        \includegraphics[width=\imgs]{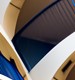} &
        \includegraphics[width=\imgs]{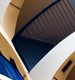} &
        \includegraphics[width=\imgs]{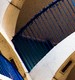} \\
        Full Image & Noisy & SwinIR~\cite{liang2021swinir} & Restormer~\cite{restormer} & Ours & Ground Truth \\
    \end{tabular}
    \end{center}
 
    \caption{Visualization results on Gaussian denoising of color images on Urban100 dataset~\cite{huang2015single_urban100}. KBNet can recover more fine textures
    }
    \label{fig:fig_gaussian}
\end{figure*}
 \begin{table*}[h]
 \caption{
    Denoising comparisons on SIDD~\cite{sidd} dataset.
    }
    \begin{center}
    \setlength{\tabcolsep}{2.pt}
    \scalebox{0.85}{
    \begin{tabular}{l |c c c c c c c c c c c c c c c c c}
    \toprule
    Method & DnCNN & MLP     & FoE   & BM3D    & WNNM    & NLM    & KSVD    & EPLL  & CBDNet  \\
     & ~\cite{DnCNN} & ~\cite{MLP} & ~\cite{FOE} & ~\cite{BM3D} & ~\cite{WNNM} & ~\cite{NLM} & ~\cite{KSVD} &~\cite{EPLL} & ~\cite{CBDNet}    \\
    \midrule
    PSNR~$\textcolor{black}{\uparrow}$ &  23.66  &   24.71  &    25.58  & 25.65  &   25.78  &   26.76  &  26.88  & 27.11  &  30.78   \\
    SSIM~$\textcolor{black}{\uparrow}$ &  0.583 &  0.641 &    0.792 &  0.685 &  0.809 &  0.699 &  0.842 &  0.870 & 0.754     \\
    \bottomrule
    \midrule
        Method & RIDNet &  VDN  & MIRNet & NBNet & Uformer & MAXIM & Restormer & NAFNet & $\mathrm{\textbf{KBNet}}_s$\\
     & ~\cite{RIDNet} & ~\cite{VDN} & ~\cite{zamir2020mirnet} & ~\cite{cheng2021nbnet} & \cite{wang2021uformer} & \cite{tu2022maxim} & \cite{restormer}  &\cite{chen2022nafnet} & (Ours) \\
    \midrule
    PSNR~$\textcolor{black}{\uparrow}$ &    38.71 & 39.28 & 39.72 & 39.75 & 39.89 & 39.96 & 40.02 & 40.30 & \textbf{40.35}\\
    SSIM~$\textcolor{black}{\uparrow}$ &      0.914 &  0.909  &  0.959 & 0.959 & 0.960 & 0.960 & 0.960 & 0.962 & \textbf{0.972} \\
    MACs~$\textcolor{black}{\downarrow}$ & 89 & - &786 & 88.8& 89.5 & 169.5 & 140 & 65 & 57.8
    \\
    \bottomrule
    \end{tabular}}
    \end{center}
    
    \label{table:sidd}
\end{table*}

\subsection{Gaussian Denoising Results}
Color and gray image denoising for Gaussian noise is widely used as benchmarks for evaluating the denoising methods. We follow the previous methods~\cite{restormer} to show color image denoising results on CBSD68~\cite{martin2001database_bsd}, Kodak24~\cite{kodak}, 
 McMaster~\cite{zhang2011color_mcmaster}, and Urban100~\cite{huang2015single_urban100} datasets. For gray image denoising, we use Set12~\cite{DnCNN}, BSD68~\cite{martin2001database_bsd}, and Urban100~\cite{huang2015single_urban100} as the testing datasets.
We train our Gaussian denoising models $\mathrm{{KBNet}}_s$ on the ImageNet validation dataset over 3 noise levels ($\sigma \in \{15, 25, 50\}$) for both color and gray images.

The results of Gaussian denoising on color images are shown in \cref{tab:c_gaussian}. 
Our method outperforms the previous state-of-the-art Restormer, but only requires half of its computational cost. 
It is also worth noticing that we do not use progressive patch sampling to improve the training performance as Restormer~\cite{restormer}. 
Compared with the window-based transformer SwinIR~\cite{liang2021swinir}, we train KBNet for 300k iterations, which is much fewer than 1.6M iterations used in SwinIR~\cite{liang2021swinir}.
The performance-efficiency comparisons with previous methods are shown in \cref{fig:trade_g}. Our KBNet achieves state-of-the-art results while requiring half of the computational cost.
Some visual results can be found in \cref{fig:fig_gaussian}. Thanks to the pixel-wise adaptive aggregation, KBNet can recover more textures even for some very thin edges.

\setlength{\tabcolsep}{2pt}
\def \imgl {.265\linewidth}
\def \imgs {0.12\linewidth}
\begin{figure*}[!t]
    \small
    \footnotesize
    \begin{center}
        \begin{tabular}{@{} c c @{}}
          \begin{tabular}{@{} c @{}}
               \includegraphics[width=\imgl]{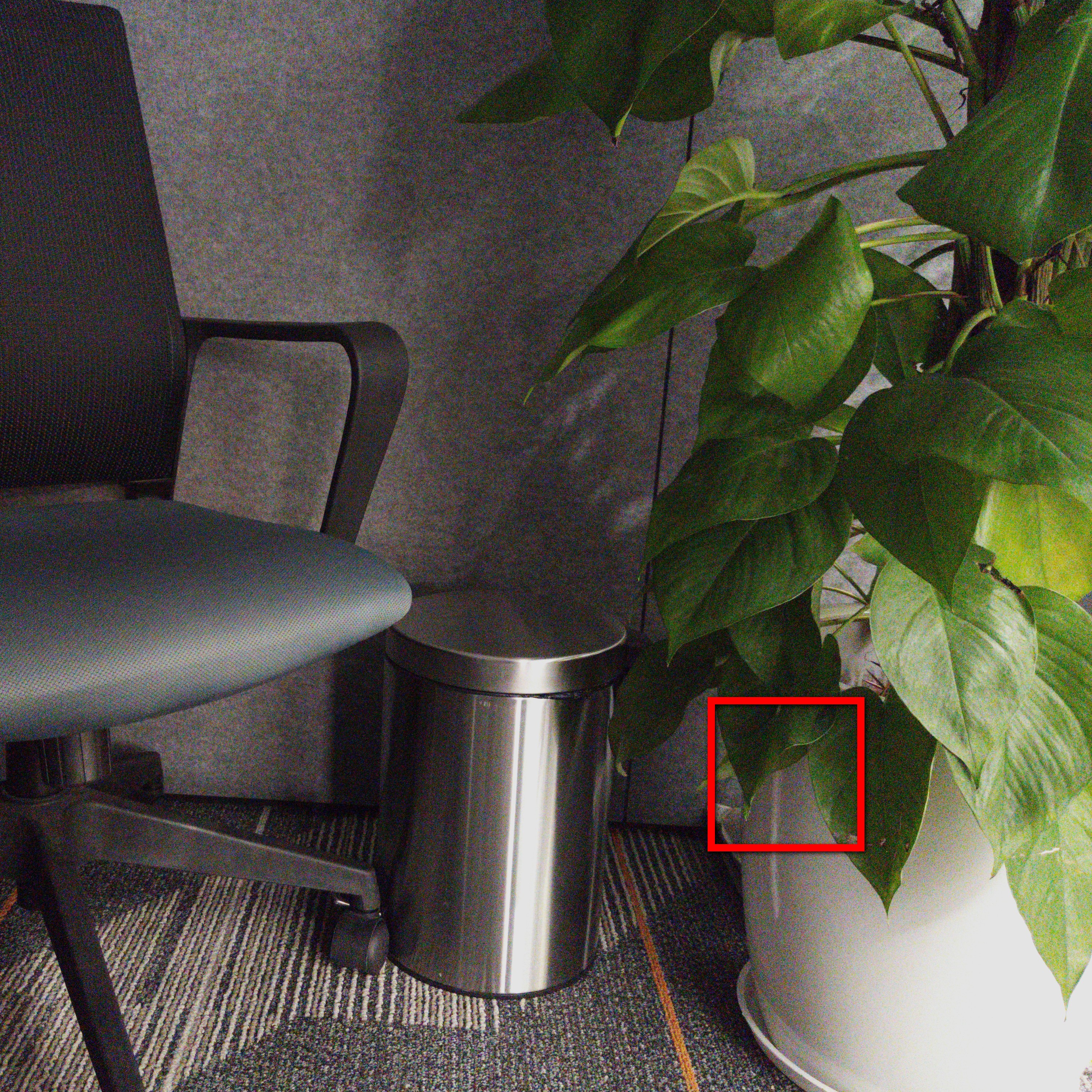} \\
               
               Full Image \\
          \end{tabular} & 
          \begin{tabular}{@{} c c c c @{}}
                \includegraphics[width=\imgs]{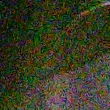} & 
                \includegraphics[width=\imgs]{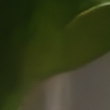} &
                \includegraphics[width=\imgs]{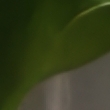} &
                \includegraphics[width=\imgs]{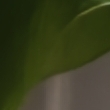} \\
                Noisy & RIDNet~\cite{RIDNet} & MIRNet~\cite{zamir2020mirnet} & Uformer~\cite{wang2021uformer}\\
                \includegraphics[width=\imgs]{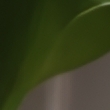} &
                \includegraphics[width=\imgs]{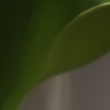} &
                \includegraphics[width=\imgs]{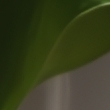} &
                \includegraphics[width=\imgs]{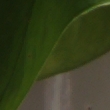} \\
                NAFNet~\cite{chen2022nafnet} & Restormer~\cite{restormer} & KBNet(Ours) & GT\\
          \end{tabular}
        \end{tabular}

    \end{center}
    \caption{Visualization of denoising results on SenseNoise dataset~\cite{zhang2021IDR}. Our method produces clearer edges and more faithful colors.
    }
    \label{fig:fig_sensenoise}
\end{figure*}
 \begin{table*}[t]
 \caption{
    Denoising comparisons on SenseNoise~\cite{zhang2021IDR} dataset.}
    \begin{center}
    \setlength{\tabcolsep}{2.pt}
    \scalebox{0.85}{
    \begin{tabular}{l| c c c c c c c c c c c c c c c c c c}
    \toprule
        Method  &  DnCNN &
     RIDNet & MIRNet & MPRNet & Uformer & Restormer & NAFNet & $\mathrm{\textbf{KBNet}}_s$ & $\mathrm{\textbf{KBNet}}_l$\\
      &~\cite{DnCNN} & ~\cite{RIDNet}  & ~\cite{zamir2020mirnet} & ~\cite{Zamir2021MPRNet} &  \cite{wang2021uformer} & \cite{restormer} & \cite{chen2022nafnet} & (Ours)& (Ours)\\
    \midrule
    PSNR~$\textcolor{black}{\uparrow}$  & 34.06 & 34.88 &
    35.30 & 35.43 & 35.43 & 35.52 & 35.55 & {35.60}& \textbf{35.69}\\  
    SSIM~$\textcolor{black}{\uparrow}$  & 0.904 & 0.915 &
    0.919 & 0.922  & 0.920& 0.924 & 0.923 & {0.924}& \textbf{0.924}\\
    MACs~$\textcolor{black}{\downarrow}$ & 37 & 89 & 130 & 120 & 90 & 80 & 65 & 57.8 & 104\\
    
    \bottomrule
    \end{tabular}}
    \end{center}
    
    \label{table:sensenoise}
\end{table*}

For Gaussian denoising of gray images, as shown in \cref{tab:g_gaussian}, KBNet shows consistent performance as the color image denoising. It outperforms the previous state-of-the-art method Restormer~\cite{restormer} slightly, but only uses less than half of its MACs.

\begin{figure}[t]
    \centering
    \includegraphics[width=0.95\linewidth]{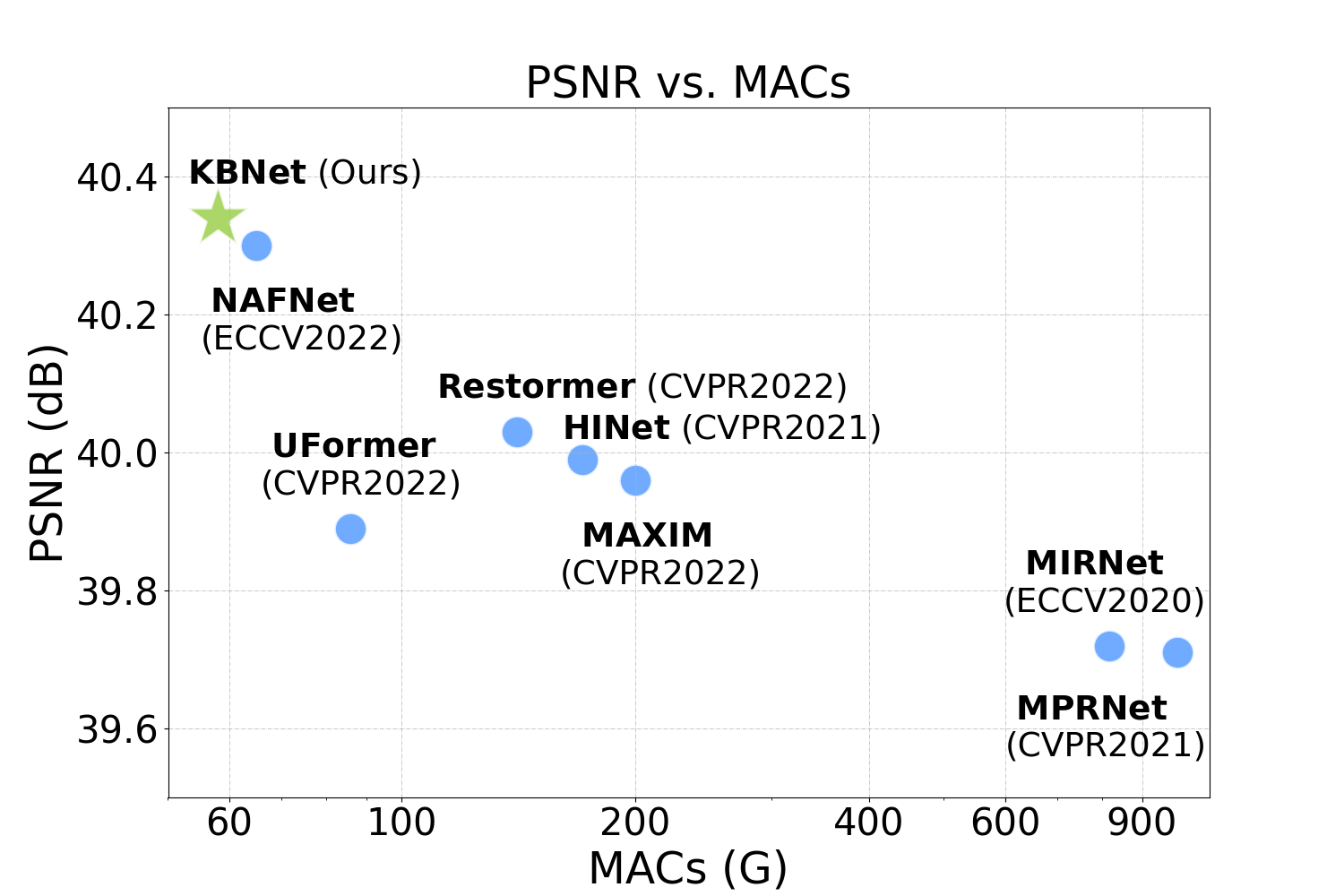}
    \caption{PSNR v.s MACs of different methods on real-world image denoising on SIDD dataset~\cite{sidd}.}
    \label{fig:trade_sidd}
\end{figure}
\begin{figure}[t]
    \centering
    \includegraphics[width=0.95\linewidth]{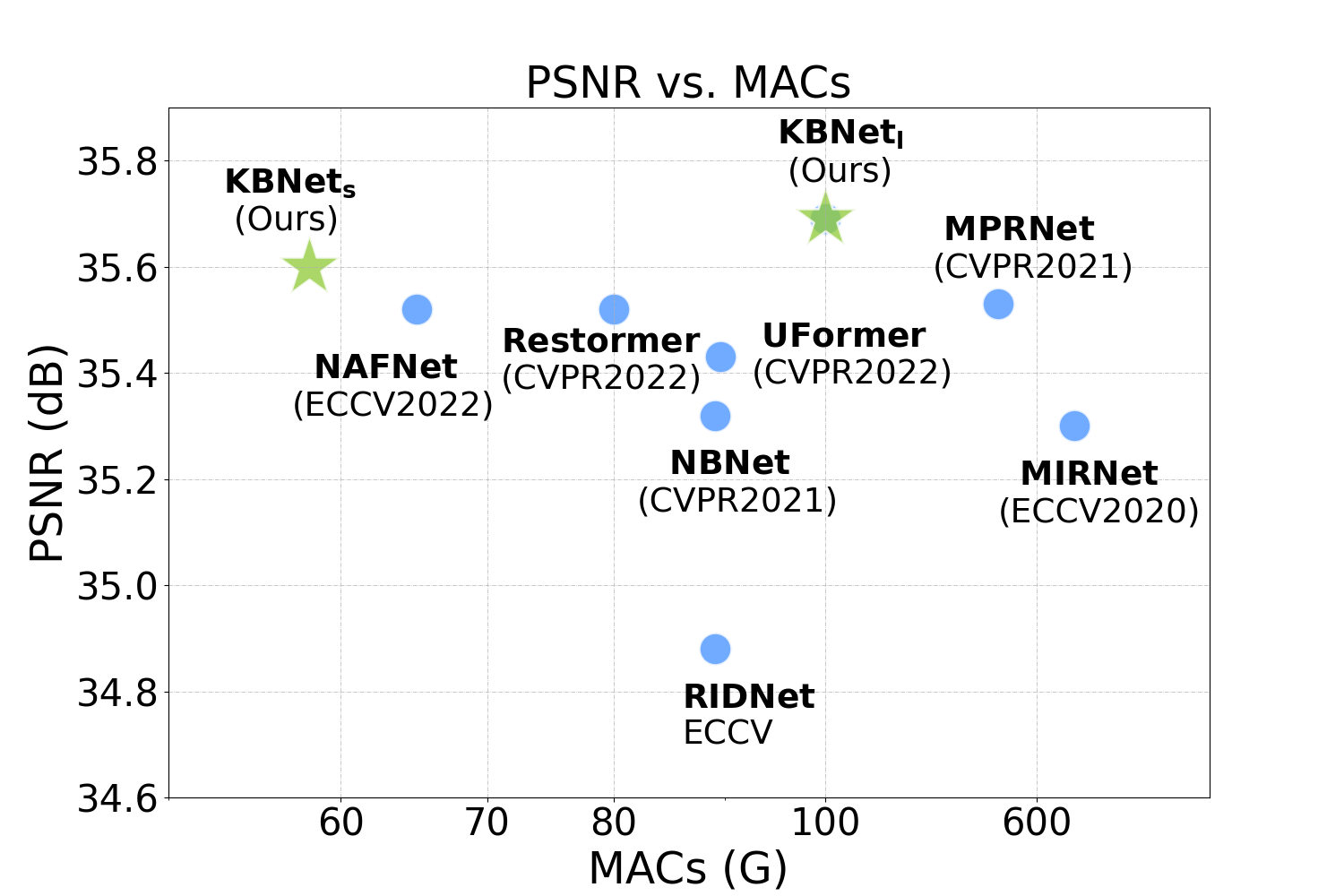}
    \caption{PSNR  v.s MACs of different methods on real-world denoising SenseNoie dataset~\cite{zhang2021IDR}.}
    \label{fig:trade_sensenoise}
\end{figure}

\subsection{Raw Image Denoising Results}
For real-world denoising, we conduct experiments on SIDD dataset~\cite{sidd} and SenseNoise dataset~\cite{zhang2021IDR} to evaluate our method on both indoor and outdoor scenes.
Besides the $\mathrm{{KBNet}}_s$, we also provide a heavier model $\mathrm{{KBNet}}_l$ that adopts the FFN design of in Restormer~\cite{restormer} to perform position-wise non-linear transformation on the SenseNoise dataset.

\noindent\textbf{SIDD:}~
The SIDD dataset is collected on indoor scenes. Five smartphones are used to capture scenes at different noise levels. SIDD contains 320 image pairs for training and $1,280$ for validation.
As shown in \cref{table:sidd}, KBNet achieves state-of-the-art results on  SIDD dataset. It outperforms very recent transformer-based methods including Restormer~\cite{restormer}, Uformer~\cite{wang2021uformer}, MAXIM~\cite{tu2022maxim} and CNN-based methods NAFNet~\cite{chen2022nafnet} with fewer MACs. \cref{fig:trade_sidd} shows the performance-efficiency comparisons of our method. KBNet achieves the best trade-offs.

\begin{table*}[!t]
\begin{center}
\caption{Comparison of defocus deblurring results on DPDD testset~\cite{abdullah2020dpdd} containing 37 indoor and 39 outdoor scenes. }
\label{table:dpdeblurring}
\scalebox{0.85}{
\begin{tabular}{l | c c c c | c c c c | c }
\toprule
   & \multicolumn{4}{c|}{\textbf{Indoor Scenes}} & \multicolumn{4}{c|}{\textbf{Outdoor Scenes}}\\
\cline{2-9}
   \textbf{Method} & PSNR~$\textcolor{black}{\uparrow}$ & SSIM~$\textcolor{black}{\uparrow}$& MAE~$\textcolor{black}{\downarrow}$ & LPIPS~$\textcolor{black}{\downarrow}$  & PSNR~$\textcolor{black}{\uparrow}$ & SSIM~$\textcolor{black}{\uparrow}$& MAE~$\textcolor{black}{\downarrow}$ & LPIPS~$\textcolor{black}{\downarrow}$ & \textbf{MACs}~$\textcolor{black}{\downarrow}$ \\
\midrule
EBDB~\cite{karaali2017edge_EBDB} & 25.77 & 0.772 & 0.040 & 0.297 & 21.25 & 0.599 & 0.058 & 0.373  &-\\
DMENet~\cite{lee2019deep_dmenet}  & 25.50 & 0.788 & 0.038 & 0.298 & 21.43 & 0.644 & 0.063 & 0.397 & 1173 \\
JNB~\cite{shi2015just_jnb} & 26.73 & 0.828 & 0.031 & 0.273 & 21.10 & 0.608 & 0.064 & 0.355  &-\\
DPDNet~\cite{abdullah2020dpdd} &26.54 & 0.816 & 0.031 & 0.239 & 22.25 & 0.682 & 0.056 & 0.313 & 991G \\
KPAC~\cite{son2021single_kpac} & 27.97 & 0.852 & 0.026 & 0.182 & 22.62 & 0.701 & 0.053 & 0.269 &- \\
IFAN~\cite{Lee_2021_CVPRifan} & {28.11}  & {0.861}  & {0.026} & {0.179}  & {22.76}  & {0.720} & {0.052}  & {0.254} & 363G 
\\
{Restormer}~\cite{restormer}& {28.87}  & {0.882}  & {0.025} & {0.145} & {23.24}  & {0.743}  & {0.050} & {0.209}  & 141G \\
\midrule
 $\mathrm{\textbf{KBNet}}_s$ & 28.42& 0.872 & 0.026 & 0.159 & 23.10 & 0.736 & 0.050 & 0.233 & 69G \\
$\mathrm{\textbf{KBNet}}_l$ & \textbf{28.89} & \textbf{0.883} & \textbf{0.024} & \textbf{0.143} & \textbf{23.32} & \textbf{0.749} & \textbf{0.049} & \textbf{0.205} & 108G \\
\bottomrule
\end{tabular}}
\end{center}
\end{table*}

\setlength{\tabcolsep}{1pt}
\begin{figure}[!t]
    \footnotesize
    \begin{center}
    \begin{tabular}{cccc}
        \includegraphics[width=.245\linewidth]{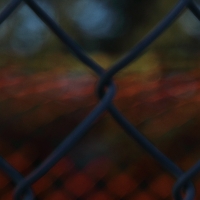} &
        \includegraphics[width=.245\linewidth]{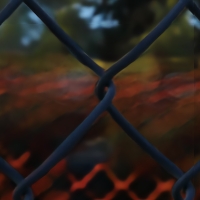} &
        \includegraphics[width=.245\linewidth]{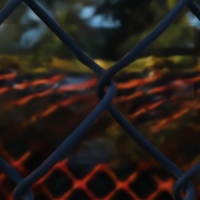} &
        \includegraphics[width=.245\linewidth]{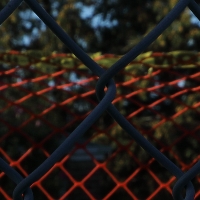} \\
        \includegraphics[width=.245\linewidth]{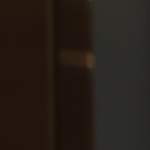} &
        \includegraphics[width=.245\linewidth]{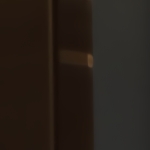} &
        \includegraphics[width=.245\linewidth]{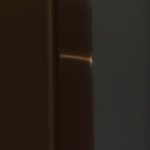} &
        \includegraphics[width=.245\linewidth]{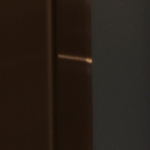} \\
        Rainy & Restormer~\cite{restormer} & Ours & GT \\
    \end{tabular}
    \end{center}
    \caption{Visualization results of defocus deblurring.}
    \label{fig:derain}
\end{figure}

\noindent\textbf{SenseNoise:}~
SenseNoise dataset contains 500 diverse scenes, where each image is of high resolution (\eg 4000 $\times$ 3000). It contains both indoor and outdoor scenes with high-quality ground truth.
We train the existing methods on the SenseNoise dataset~\cite{zhang2021IDR} under the same training setting of NAFNet~\cite{chen2022nafnet} but use 100k iterations. Since some of the models are too heavy, we scale down their channel numbers for fast training.
The performance and MACs are reported in \cref{table:sensenoise}. 
Our method not only outperforms all other methods but also achieves the best performance-efficiency trade-offs as shown in \cref{fig:trade_sensenoise}.
Some visualizations are shown in \cref{fig:fig_sensenoise}.
KBNet produces sharper edges and recovers more vivid colors than previous methods.

\subsection{Deraining and Defocus results}
To demonstrate the generalization and effectiveness of our KBNet, we follow the state-of-the-art image restoration method Restormer~\cite{restormer} to conduct experiments on deraining and defocus deblurring. 
The channels of our MFF module are adjusted to make sure our model uses fewer MACs than Restormer~\cite{restormer}. The training settings are kept the same as the Restormer~\cite{restormer}.

\begin{table*}
\begin{center}
\caption{\small Image deraining results.}
\label{table:deraining}

\setlength{\tabcolsep}{9.5pt}
\scalebox{0.85}{
\begin{tabular}{l c c c c | cc | c}
\toprule
  & \multicolumn{2}{c}{\textbf{Test2800}~\cite{fu2017removing}}&\multicolumn{2}{c|}{\textbf{Test1200}~\cite{zhang2018density}}\\
 \textbf{Method} &   PSNR~$\textcolor{black}{\uparrow}$ & SSIM~$\textcolor{black}{\uparrow}$ & PSNR~$\textcolor{black}{\uparrow}$ & SSIM~$\textcolor{black}{\uparrow}$ & \textbf{MACs}~$\textcolor{black}{\downarrow}$\\
\midrule
DerainNet~\cite{fu2017clearing} &  24.31  & 0.861  & 23.38  & 0.835     &-\\
SEMI~\cite{wei2019semi} & 24.43&0.782& 26.05&0.822 &-\\
DIDMDN~\cite{zhang2018density} &  28.13&0.867& 29.65&0.901 &-\\
UMRL~\cite{yasarla2019uncertainty} & 29.97&0.905& 30.55&0.910 &-\\
RESCAN~\cite{li2018recurrent} &  31.29&0.904& 30.51&0.882 &-\\
PreNet~\cite{ren2019progressive} & 31.75&0.916& 31.36&0.911& 66.2G \\
MSPFN~\cite{mspfn2020}  & 32.82 & 0.930 & 32.39 & 0.916 &-\\
MPRNet~\cite{Zamir_2021_CVPR_mprnet}  &  {33.64} & {0.938} & 32.91 & 0.916 &  1.4T \\
SPAIR~\cite{purohit2021spatially_spair} & 33.34 & 0.936 & {33.04} &{0.922}  & - \\
{Restormer}~\cite{restormer} &  {34.18} & {0.944} & {33.19} & {0.926}  & 141G \\
\midrule
$\mathrm{\textbf{KBNet}}_s$ & 33.10 & 0938 & 32.29 & 0.912 & 69G \\
$\mathrm{\textbf{KBNet}}_l$ &  \textbf{34.19} & \textbf{0.944} & \textbf{33.82} & \textbf{0.931}  & 108G \\
\bottomrule
\end{tabular}}
\end{center}
\end{table*}

 \begin{table}[t]
 \caption{
    Ablation studies on the dynamic spatial aggregation design choices.}
    \begin{center}
    \scalebox{1.}{
    \begin{tabular}{l| c c c c}
    \toprule
        Method & PSNR &  SSIM & MACs \\
        
    \midrule
    Dynamic conv~\cite{AttOverConv20} & 29.31 & 0.881  & 12G \\
    Kernel prediction module~\cite{kpn}  & 29.29 & 0.881 & 55.1G \\
    Shifted window attention~\cite{wang2021uformer} & 29.33 & 0.882  & 21.9G \\
    \midrule
    KBA w/ softmax &  29.40 & 0.883  & 15.3G\\
    KBA (Ours) & 29.47 & 0.884  & 15.3G  \\
    
    \bottomrule
    \end{tabular}}
    \end{center}
    
    \label{table:ab_soft}
\end{table}

 \begin{table}[t]
 \caption{The influence of the kernel basis number.}
    \begin{center}
    \scalebox{1.}{
    \begin{tabular}{l| c c c c}
    \toprule
        Kernel Bases Number & PSNR   \\
    \midrule
    $N = 4$ & 29.36  \\  
    $N = 8$ & 29.41  \\  
    $N = 16$ & 29.44 \\  
    $N = 32$ & 29.47 \\  
    $N = 64$ & 29.51 \\ 
    $N = 128$ & 29.54 \\ 
    \bottomrule
    \end{tabular}}
    \end{center}
    
    \label{tab:nbases}
\end{table}

 \begin{table}[t]
 \caption{The effectiveness of different branches in MFF block.}
    \begin{center}
    \scalebox{1.}{
    \begin{tabular}{l| c c c c}
    \toprule
        Method & PSNR   \\
    \midrule
    DW$3\times3$  & 29.12  \\  
    DW$3\times3$ + CA & 29.22  \\  
    DW$3\times3$ + CA + KBA& 29.47     \\    
    \bottomrule
    \end{tabular}}
    \end{center}
    
    \label{table:ab_branch}
\end{table}
\setlength{\tabcolsep}{1pt}
\begin{figure}[!t]
    \footnotesize
    \begin{center}
    \begin{tabular}{cccc}
        \includegraphics[width=.245\linewidth]{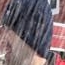} &
        \includegraphics[width=.245\linewidth]{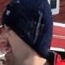} &
        \includegraphics[width=.245\linewidth]{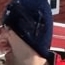} &
        \includegraphics[width=.245\linewidth]{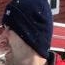} \\
        \includegraphics[width=.245\linewidth]{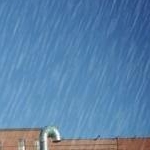} &
        \includegraphics[width=.245\linewidth]{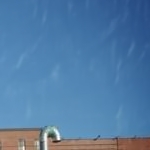} &
        \includegraphics[width=.245\linewidth]{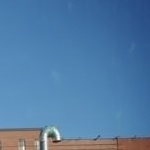} &
        \includegraphics[width=.245\linewidth]{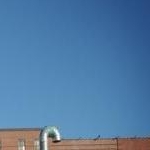} \\
        Rainy & Restormer~\cite{restormer} & Ours & GT \\
    \end{tabular}
    \end{center}
    \caption{Visualization results of deraining.}
    \label{fig:derain}
\end{figure}

\noindent\textbf{Deraining}. The largest two datasets (Test2800 and Test1200) are used for testing deraining performance. 
Results in \cref{table:deraining} indicate that our KBNet has a good generalization on deraining. 
$\mathrm{KBNet}_l$ outperforms Restormer~\cite{restormer} using only $76\%$ of its MACs.
On the Test1200 dataset, $\mathrm{KBNet}_l$ produces more than 0.5dB improvement.
Some visualization results can be found in \cref{fig:derain}.

\noindent\textbf{Defocus Deblurring}. As shown in \cref{table:dpdeblurring}, we test our model on both indoor and outdoor scenes for deblurring. 
$\mathrm{KBNet}_s$ outperforms most previous methods using only 69G MACs.
$\mathrm{KBNet}_l$ outperforms previous state-of-the-art Restormer~\cite{restormer}  while having $24\%$ fewer MACs.
Some visualization results are shown in \cref{fig:derain}.

\subsection{Ablation Studies}
We conduct extensive ablation studies to validate the effectiveness of components of our method and compare it with existing methods.
All ablation studies are conducted on Gaussian denoising with noise level $\sigma=50$. 
We train models for 100k iterations. Other training settings are kept the same as the main experiment on the Gaussian denoising of color images.

\noindent\textbf{Comparison with dynamic spatial aggregation methods.}~
We compare our method with existing dynamic aggregation solutions, including the window-based self-attention~\cite{wang2021uformer}, dynamic convolution~\cite{AttOverConv20}, and kernel prediction module~\cite{kpn,Wang2019Carafe} to replace our KBA module in the proposed MFF blocks. 
As shown in \cref{table:ab_soft}, directly adopting the kernel prediction module in ~\cite{kpn} requires heavy computational cost. While the dynamic convolution~\cite{AttOverConv20} predicts spatial-invariant dynamic kernels, it slightly outperforms the kernel prediction module. 
This demonstrates that the kernel prediction module is difficult to optimize as it requires predicting a large number of kernel weights directly. Most existing works~\cite{kpn,mckpn,xia2020bpn} need to adopt a heavy computational branch to realize the kernel prediction.
The shifted window attention requires more MACs while only improving the performance marginally than the dynamic convolution.
Our method is more lightweight and improves performance significantly.

\noindent\textbf{Using softmax in KBA module.}~
To linearly combine the kernel bases, a natural choice is to use the softmax function to normalize the kernel fusion coefficients. 
However, we find that using the softmax would hinder the performance as shown in \cref{table:ab_soft}. 
Using the softmax function may encourage the fused kernel more focus on a specific kernel basis and reduces the diversity of the fused kernel weights.

\noindent\textbf{The impact of the number of kernel bases.}~
We also validate the influence of the number of kernel bases. As shown in \cref{tab:nbases}, more kernel bases bring consistent performance improvements since it captures more image patterns and increases the diversity of the spatial information aggregation. 
In our experiments, we select $N=32$ for a better performance-efficiency trade-off.

\noindent\textbf{The impact of different branches in MFF block.}~
As shown in \cref{table:ab_branch}, a single $3 \times 3$ depthwise convolution branch produces 29.12 dB on Gaussian denoising. Adding the channel attention branch and KBA module branch successively leads to 0.1dB and 0.25dB respectively. KBA module brings the largest improvement, which indicates the importance of the proposed pixel adaptive spatial information processing.

\setlength{\tabcolsep}{1pt}
\begin{figure}[!t]
    \footnotesize
    \begin{center}
    \begin{tabular}{cccc}
        \includegraphics[width=.245\linewidth]{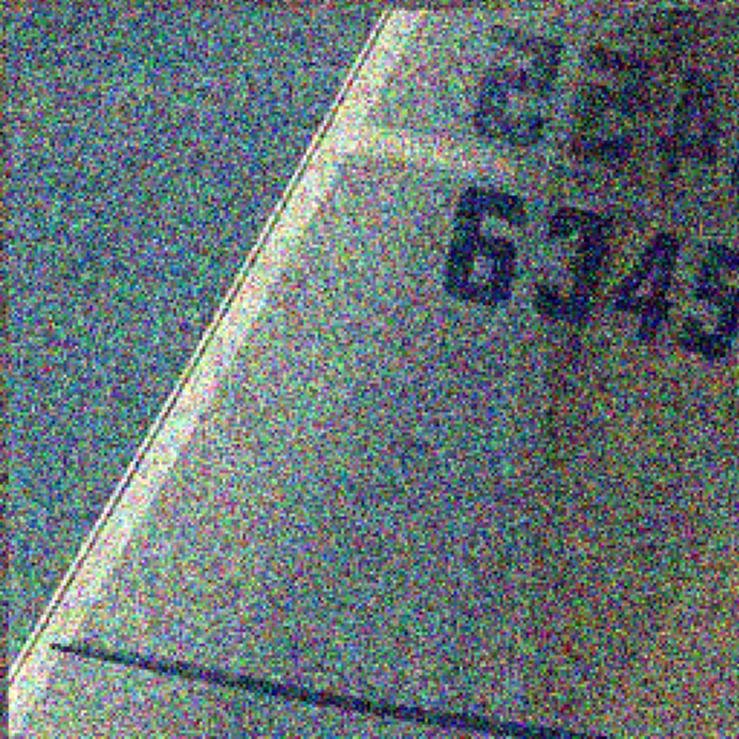} &
        \includegraphics[width=.245\linewidth]{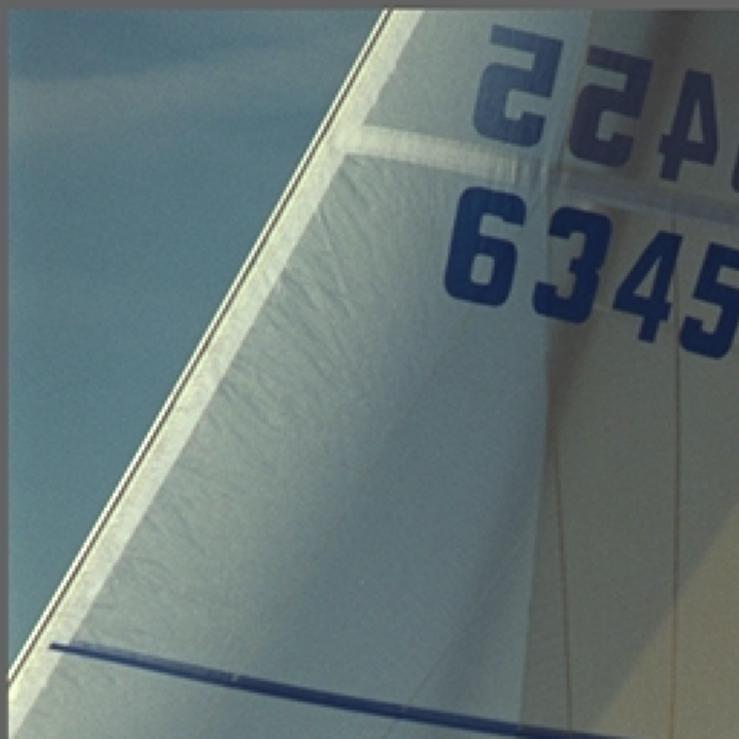} &
        \includegraphics[width=.245\linewidth]{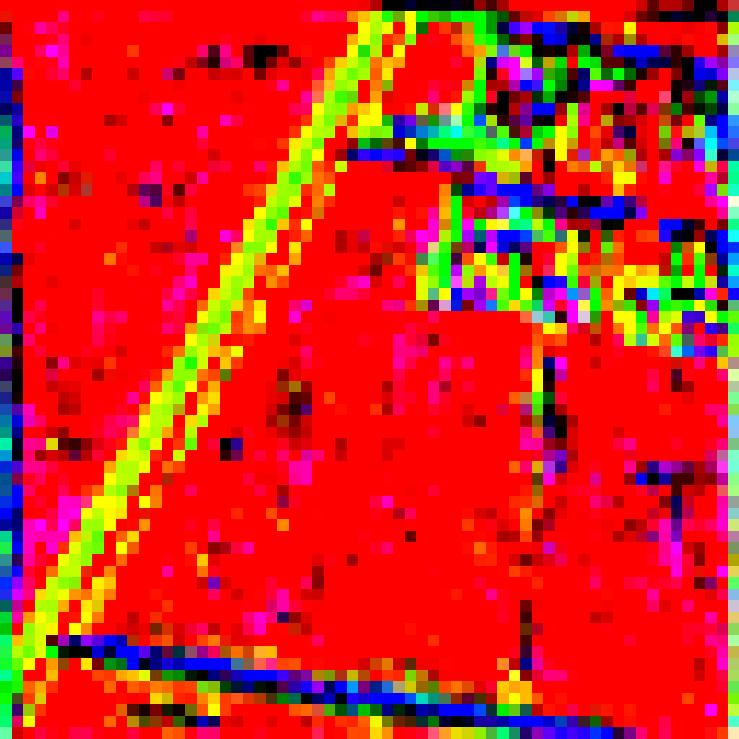} &
        \includegraphics[width=.245\linewidth]{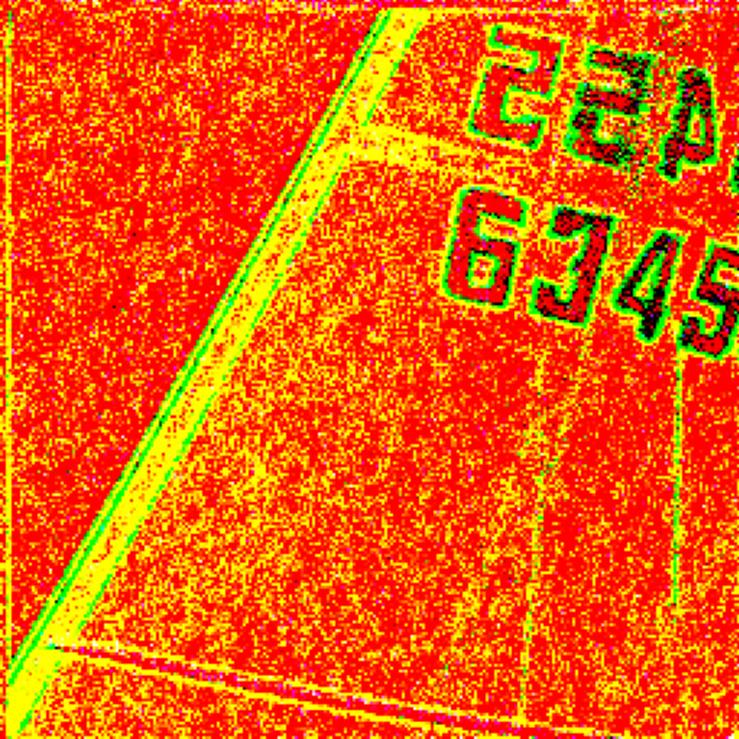} \\
        \includegraphics[width=.245\linewidth]{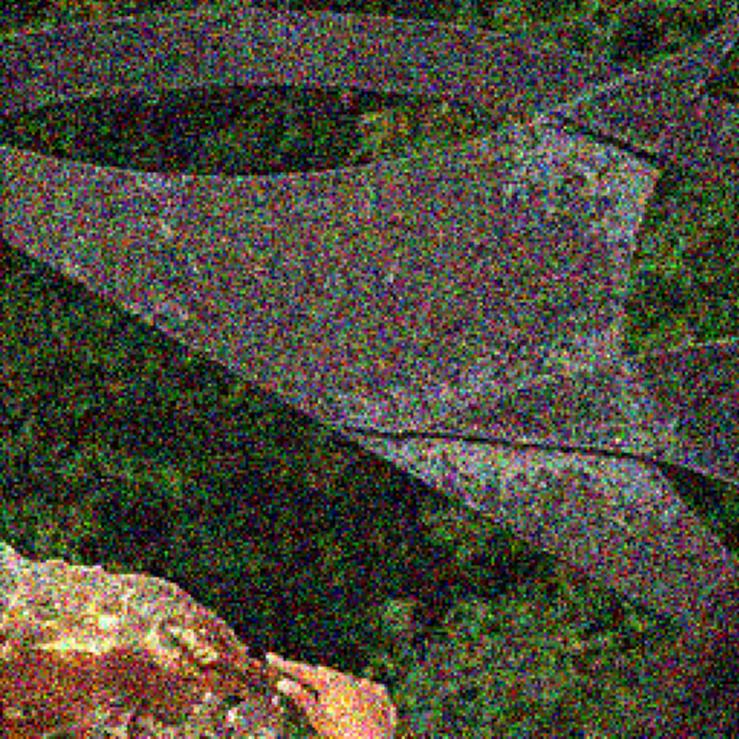} &
        \includegraphics[width=.245\linewidth]{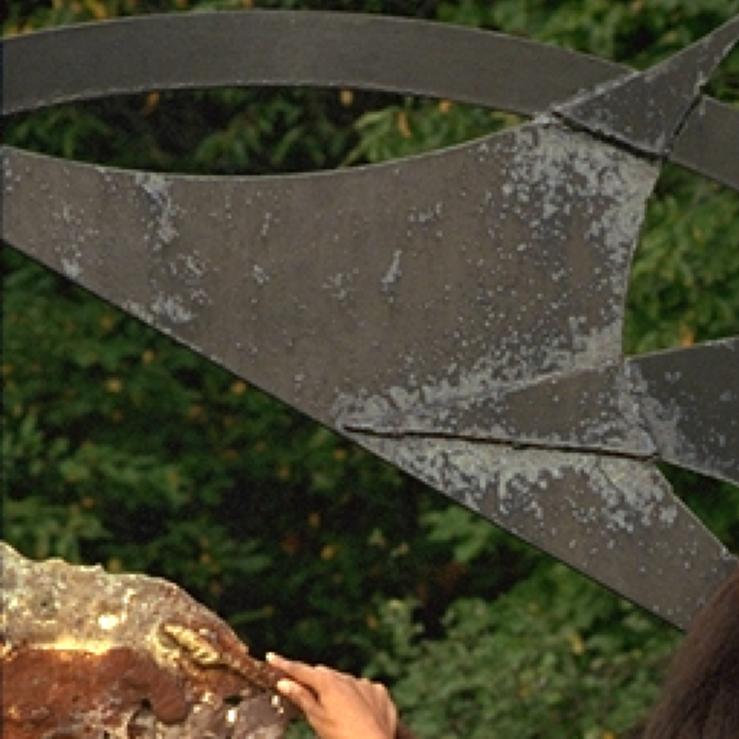} &
        \includegraphics[width=.245\linewidth]{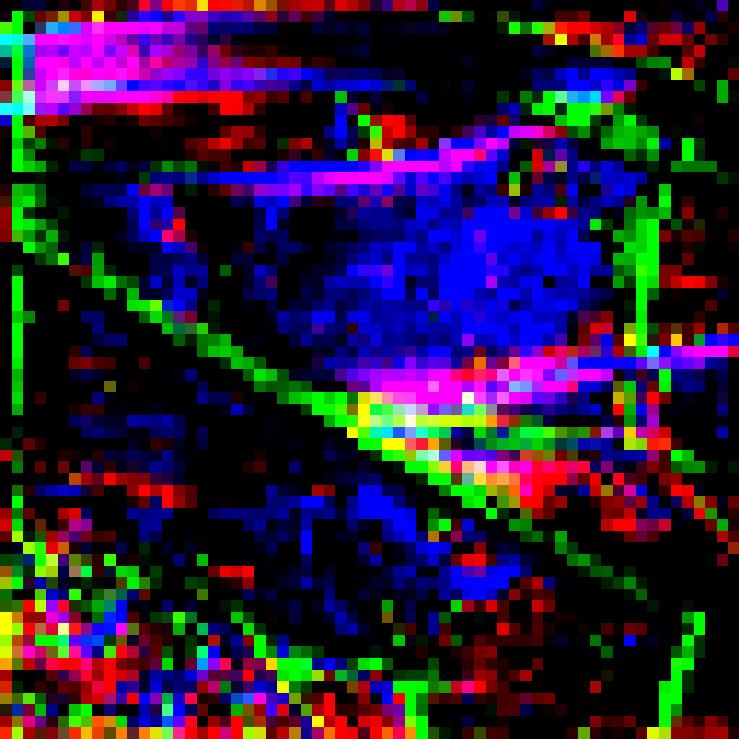} &
        \includegraphics[width=.245\linewidth]{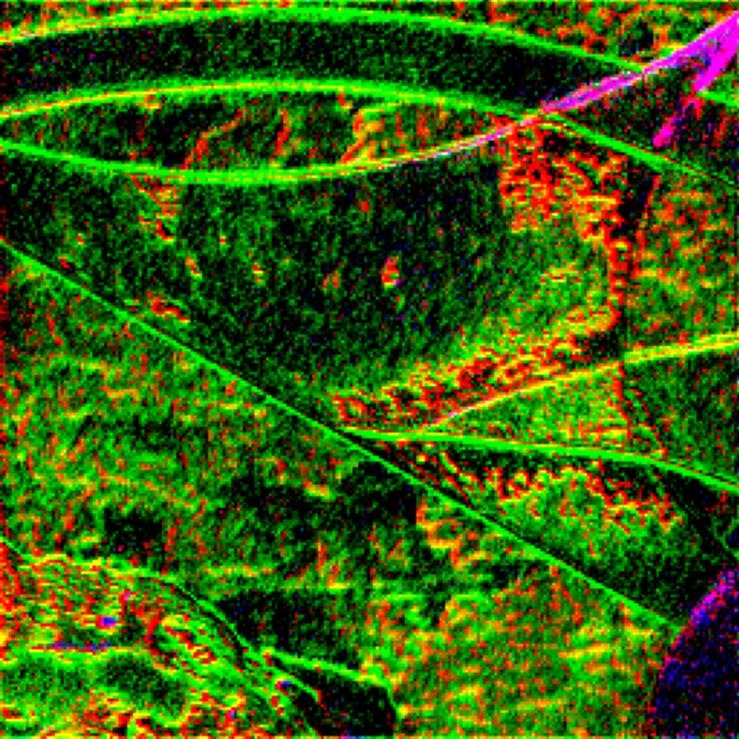} \\
        Noisy & GT & $64 \times 64$ & $256 \times 256$ \\
    \end{tabular}
    \end{center}
    \caption{Visualization of kernel indices in different stages. The fusion coefficient map is transformed into a 3D RGB map by random projection.}
    \label{fig:kernel}
\end{figure}
\noindent\textbf{Visulization of kernel indices.}~
We visualize the kernel indices of different regions in
different stages of our KBNet. 
We project the fusion coefficients to a 3D space by random projection and visualize them as an RGB map. 
As shown in \cref{fig:kernel}, similar color indicates that the pixels fuse kernel bases having similar patterns. 
KBNet can identify different regions and share similar kernel bases for similar textural or plain areas.
Different kernels are learned to be responsible for different regions, and they can be optimized jointly during the training. 

\section{Conclusion}
In this paper, we introduce a kernel basis network (KBNet) for image restoration. 
The key designs of our KBNet are kernel basis attention (KBA) module and Multi-axis  Feature Fusion (MFF) block. KBA module adopts the learnable kernel bases to model the local image patterns and fuse kernel bases linearly to aggregate the spatial information efficiently. 
Besides, the MFF block aims to fuse diverse features from multiple axes for image denoising, which includes channel-wise, spatial-invariant, and pixel-adaptive processing.
In the experiments, KBNet achieves state-of-the-art results on popular synthetic noise datasets and two real-world noise datasets (SIDD and SenseNoise) with less computational cost. It also presents good generalizations and state-of-the-art results on deraining and defocus deblurring.



\backmatter

\bibliographystyle{plain}
\bibliography{egbib}

\end{document}